\documentclass[11pt]{article}

\usepackage[preprint]{acl}

\usepackage{times}
\usepackage{latexsym}
\usepackage[T1]{fontenc}
\usepackage[utf8]{inputenc}
\usepackage{microtype}
\usepackage{inconsolata}
\usepackage{graphicx}
\usepackage{booktabs}
\usepackage{amsmath}
\usepackage{amssymb}
\usepackage{array}
\usepackage{enumitem}
\usepackage{xcolor}

\newcommand{\dsa}{\emph{The Dark Eye}}

\title{Investigating the Role of Reasoning-Language Alignment in\\ Monolingual Retrieval-Augmented Generation}

\author{
    Oliver Hauck$^{*1}$\quad~
    Mario Sanz-Guerrero$^{*1}$\quad~
    Katharina von der Wense$^{1,2}$\\
    $^1$Johannes Gutenberg University Mainz, Germany \\
    $^2$University of Colorado Boulder, USA \\
    \texttt{\href{mailto:msanz@uni-mainz.de}{msanz@uni-mainz.de}}
}

\begin{document}
\maketitle

\renewcommand{\thefootnote}{\fnsymbol{footnote}}
\footnotetext[1]{Equal contribution.}
\renewcommand{\thefootnote}{\arabic{footnote}}

\begin{abstract}
Reasoning traces improve large language models (LLMs), but current models are trained to reason mostly in English. It has been shown that forcing a model to reason in another language degrades accuracy, even when the reasoning language matches the language of the prompt -- but only for a setting where the model reasons over a short prompt. Here, we ask whether the same holds for retrieval-augmented generation (RAG), where the model must read and integrate a large amount of retrieved evidence in the target language. To study this, we build a fully monolingual German RAG question-answering testbed over the fictional world of the tabletop role-playing game \dsa{}, a domain that is richly documented in German but too niche for the model to answer from memory, so that it has to rely on retrieval. Varying the forced reasoning language of an agentic RAG system on this testbed, we find that aligning the reasoning language with the language of the query and the retrieved documents helps. Forced German reasoning outperforms forced French, although the model benchmarks higher in French, so the benefit comes from alignment and not from language proficiency. The advantage grows when the retrieved context is richer and structure-aware. However, forced German only reaches the level of the model's native, unconstrained English reasoning without surpassing it, showing that 
native multilingual reasoning is needed. We publicly release the testbed and QA benchmark.
\end{abstract}

\section{Introduction}

Large language models (LLMs) increasingly produce an explicit \emph{reasoning trace} before their final answer. This improves accuracy on many tasks and lets users inspect how the model reached its answer \citep{deepseekai2025deepseekr1,xu2025towards,chen2025reasoningerasurveylong}. Current reasoning models, however, are trained to reason mostly in English (or, for some models, Chinese), which limits their usefulness for speakers of other languages.

\begin{figure}[t]
\centering
\includegraphics[width=\columnwidth]{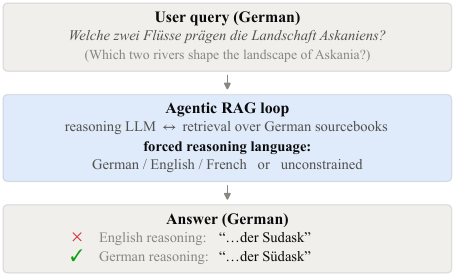}
\caption{Our setting. An agentic RAG system answers German questions using retrieved German sourcebook passages; we control the language the model reasons in. In this real example, English reasoning drops the umlaut of the river \emph{Südask}, while German reasoning keeps it.}
\label{fig:system}
\end{figure}
\citet{qi2025reasoning} show that this bias is hard to remove at inference time. Forcing a model to reason in a chosen language degrades its accuracy, and the drop remains even when the forced reasoning language is the same as the language of the prompt. Their study, however, evaluates the model on short prompts, where the model reasons over the question alone. This leaves open an important case. Many real deployments for non-English users rely on \emph{retrieval-augmented generation} (RAG), where the model is given long passages retrieved from a knowledge base and must read and combine them to answer \citep{lewis2020retrieval}. In this case, most of the text the model reasons over is retrieved evidence in the target language, not just the question.
This motivates our research question: \emph{in a monolingual non-English RAG setting, where the query and the retrieved documents are all in the target language, does forcing the model to reason in that language still hurt, or does aligning the reasoning language with the evidence help?}

There is a reason to expect that alignment could help here more than in the prompt-only case. For an LLM, language is the only way it perceives the world, and everything it sees comes through tokens, which encode language-specific meaning \citep{wolff2011linguistic}. When the model must interpret a large amount of retrieved German text, reasoning in German may keep the reasoning trace closer to the evidence, whereas reasoning in English forces an implicit translation. Figure~\ref{fig:system} shows a concrete case. For a German query, an English reasoning trace produces an answer that drops a German umlaut (\emph{Sudask}), while a German reasoning trace keeps it (\emph{Südask}).

Testing this needs a setting that is (i) fully monolingual in a non-English language and (ii) genuinely dependent on retrieval, so that the effect we measure is about reasoning over retrieved evidence and not about knowledge the model already has. We therefore build a testbed over \dsa{} (German: \emph{Das Schwarze Auge}),\footnote{DAS SCHWARZE AUGE, AVENTURIEN, THE DARK EYE, and AVENTURIA are registered trademarks of Ulisses Spiele GmbH. They are named here only for academic and descriptive purposes.} a popular German tabletop role-playing game whose fictional world \emph{Aventuria} is described in detail across decades of German sourcebooks but is too niche for current LLMs to know from pretraining. In our experiments a model without retrieval scores near the minimum on our questions (Section~\ref{sec:results}), which confirms that the model must use the retrieved German text to answer.

On this testbed we run an agentic RAG system that uses \texttt{Qwen3-30B-A3B} \citep{qwen3technicalreport} and vary the forced reasoning language (German, English, French, unconstrained, and no reasoning) and the chunking strategy of the knowledge base. Our main findings are:

\begin{itemize}
\item Aligning the reasoning language with the language of the query and retrieved documents helps. Forced German reasoning clearly beats forced French, even though the model benchmarks \emph{higher} in French than German. The benefit therefore comes from \emph{alignment} with the data, not from language proficiency.
\item The alignment benefit grows with richer, structure-aware target-language context. Moving from plain fixed-size chunks to section-based chunks improves forced German reasoning the most and also reduces the general accuracy cost of forcing a reasoning language.
\item Forced German nonetheless does not surpass the model's native, unconstrained English reasoning. Prompt-based language control narrows the gap but does not overturn the English bias, which points to native multilingual reasoning as the real fix.
\item Reasoning itself helps only above a task-complexity threshold: disabling reasoning is best on simple single-passage questions but worst on complex multi-hop questions.
\end{itemize}

Alongside these findings we release the testbed as a contribution itself: a structure-preserving pipeline that turns richly formatted sourcebook PDFs into a hierarchical German knowledge base, and a curated German QA benchmark of 585 single-passage and 30 multi-hop questions with human evaluation. The RAG architecture itself is standard; what the testbed adds is a controlled, fully monolingual QA setting in which reasoning over target-language evidence can be studied.\footnote{\label{fn:code}Code and QA benchmark: \url{https://github.com/soliver1/reasoning-language-alignment}.}

\section{Related Work}

\paragraph{Retrieval-augmented generation.}
RAG augments an LLM with an external knowledge base at inference time by retrieving passages relevant to the query and adding them to the context \citep{lewis2020retrieval}. It reduces hallucination and adds knowledge without changing the model's parameters \citep{Huang2025SurveyHallucinations,gao2023retrieval}. A central design choice is \emph{chunking}. Fixed-size chunks are simple but can mix unrelated topics and mislead the model \citep{gao2023retrieval}. Recent work instead uses \emph{semantic} or \emph{hierarchical} chunking that respects document structure such as headings \citep{Latif2025TheCP,zhong2025semrag,nguyen2025enhancing,lu2025hichunk}. Retrieval precision is often improved with a reranking step, where a cross-encoder re-scores the top candidates \citep{glass2022re2g,oche2025systematicreview}. Our knowledge base uses a simple hierarchical (section-based) chunking scheme and a reranker; the pipeline is standard, and we use it as a controlled setting rather than as a new retrieval method.

\paragraph{Multilingual and cross-lingual RAG.}
RAG has been studied mostly in English, but a growing line of work asks how it behaves in other languages. \citet{chirkova2024multilingual} build an mRAG pipeline over 13 languages and find that generating in the user's language needs task-specific prompting, and that outputs suffer from code-switching and fluency errors. \citet{wu2024notall} separate monolingual retrieval from cross-lingual transfer and report a consistent advantage for English and for high-resource languages. Benchmarks such as MIRACL \citep{zhang2023miracl} and NoMIRACL \citep{thakur2024knowing} measure multilingual retrieval quality and robustness to irrelevant retrieved passages. \citet{ranaldi-etal-2026-multilingual} compare retrieval strategies for multilingual QA and find it best to translate the retrieved documents into a common language before generation. All of this work varies the language of the \emph{query} or the \emph{documents}. We instead hold both fixed -- query, corpus, and answer are German -- and vary only the language the model \emph{reasons} in, an axis these studies do not isolate.

\paragraph{Multilingual reasoning.}
Reasoning models generate a trace before answering, which improves accuracy on complex tasks \citep{deepseekai2025deepseekr1,xu2025towards}. Because these models reason predominantly in English, several works try to improve reasoning in other languages \citep{HuanMindMerger2024,huang2023languagescreatedequalllms}. Most relevant to us, \citet{qi2025reasoning} force the reasoning language with a prompt-hacking technique and report an accuracy drop under such control, even when the reasoning language matches the prompt language; light fine-tuning reduces but does not remove the drop. Their evaluation is prompt-only. Closer to a RAG setting, \citet{ranaldi-etal-2025-improving-multilingual} improve grounding over multilingual evidence by structuring the reasoning as an argumentative process that resolves conflicting passages; they shape \emph{how} the model reasons over retrieved evidence, whereas we vary the \emph{language} it reasons in. We ask whether the effect of \citet{qi2025reasoning} holds in a RAG setting, where the model reasons over a large amount of retrieved evidence in the target language, and we separate the effect of \emph{language alignment} from that of \emph{language proficiency}.

\paragraph{LLMs and tabletop role-playing games.}
Tabletop role-playing games (TTRPGs) are narrative-driven, which has motivated work on LLM game-master assistants \citep{zhu2023calypso,kelly2023towards} and autonomous game masters \citep{triyason2023exploring,ang2023fable}; see \citet{shyne2025computational} for a survey. A recurring problem is the model's limited knowledge of the specific game world, which leads to answers that do not match the game world \citep{gongora2024payador}. We use one such world not as an application but as a knowledge source that the model provably does not know, which is exactly what makes it a clean testbed for retrieval-grounded reasoning. Our curated knowledge base could also serve future TTRPG systems.

\section{A Monolingual RAG Testbed for \dsa{}}
\label{sec:testbed}

\dsa{} is a German TTRPG first published in 1984, set in the fictional world \emph{Aventuria} \citep{DSA5}. Its setting is documented in an unusually detailed way across decades of German sourcebooks. This combination -- rich German documentation, but a niche world absent from LLM pretraining -- is what we need: the model cannot answer from memory and must use the retrieved German text.

\subsection{Structure-Preserving Knowledge Base}
\label{sec:testbed:kb}

Our source documents are the \emph{green series} of region description books for \dsa{}. Because of the substantial manual effort described below, we use three of them (G01, G05, G08) \citep{G01,G05,G08}, each describing a different region. Together they amount to about 1.13M tokens, far beyond the context window of the model we use, so retrieval is required.

Unlike plain text, books already carry a hierarchy of chapters, sections, and subsections. Following recent hierarchical-chunking work \citep{nguyen2025enhancing,lu2025hichunk}, we keep this structure. Each chunk holds one section (a heading and its text) and is annotated with links to its parent and child chunks.

This matters because a section is often not self-contained. Figure~\ref{fig:chunk} shows a real example. A short paragraph about market stalls and taxes cannot be understood on its own, but becomes clear once the retrieved chunk carries its ancestor headings, which place it in the city of Al'Anfa. The structure is what supplies that context, and it is also what enables fine-grained citations back to a specific section and page rather than to a page alone. This is the ``German structure'' that Section~\ref{sec:results:context} shows forced German reasoning benefits from most.

\begin{figure}[t]
\centering
\small
\setlength{\fboxsep}{5pt}
\noindent\fbox{\begin{minipage}{0.93\columnwidth}
\ttfamily\footnotesize\raggedright
\# Al'Anfa -- Stadt und Imperium\\
\#\# Al'Anfa -- Die schwarze Perle des Südens\\
\#\#\# Märkte\\[2pt]
\rmfamily\itshape Auf den meisten Plätzen [\ldots] Abgabe auch geleistet hat.\\[3pt]
\rmfamily\upshape\scriptsize \textbf{id} G01-099 \quad \textbf{parent} G01-095 \quad \textbf{children} --
\end{minipage}}
\caption{A retrieved chunk from the SectionChunks knowledge base. The section text alone (\emph{``Märkte''}) does not say which city it describes; the ancestor headings carried with the chunk do. Chunk IDs let the model walk to the parent or child section.}
\label{fig:chunk}
\end{figure}

Extracting this structure from the PDFs is the main engineering effort. The books use a dense two- and three-column layout, and off-the-shelf PDF-to-text tools \citep[Docling, Marker, pdfalto;][]{Docling,marker,pdfalto} all fail to recover the reading order and section boundaries reliably. We therefore build a semi-automatic pipeline that converts each PDF to a structured representation, using pdfalto to keep block and font information, recovering words that the converter drops, and manually correcting heading levels and block order where automatic decisions are unreliable. The pipeline is described in Appendix~\ref{sec:appendix:pipeline}; it is only a way to obtain a faithful knowledge base, not a contribution we evaluate.

\subsection{QA Benchmark}
\label{sec:testbed:qa}

We build two German QA sets over this knowledge base.

\paragraph{Single-passage set (585 questions).}
In this dataset, each question is answerable from a single section. Following \citet{yuen2025automaticdataset}, we generate questions with an LLM. For each section, a generator model proposes question--answer pairs, a second \emph{critic} model rates them and returns feedback, and low-rated pairs are revised. A human reviewer then accepts, rejects, or edits the best pair per section. The final set has 585 human-checked pairs. Because it is built one section at a time, it tests retrieval and reading, but not multi-passage synthesis.

\paragraph{Multi-hop set (30 questions).}
To test synthesis across sections, we hand-write 30 harder questions that cannot be answered from a single chunk and that draw on all three books. This set is small but high-quality, and (as it needs domain expertise and multi-source evidence to judge) is evaluated by human experts rather than automatically.

\paragraph{Data and code.}
The sourcebooks are under copyright, so we cannot share the processed knowledge base itself. Instead, we release the tool that builds it from the PDFs.\footref{fn:code} Anyone who owns the original books can run this tool and obtain the same knowledge base we use, so our results are reproducible without us redistributing copyrighted text. We also release the QA benchmark and the system code.

\section{Agentic RAG System}
\label{sec:system}

\begin{figure*}[t]
\centering
\includegraphics[width=\textwidth]{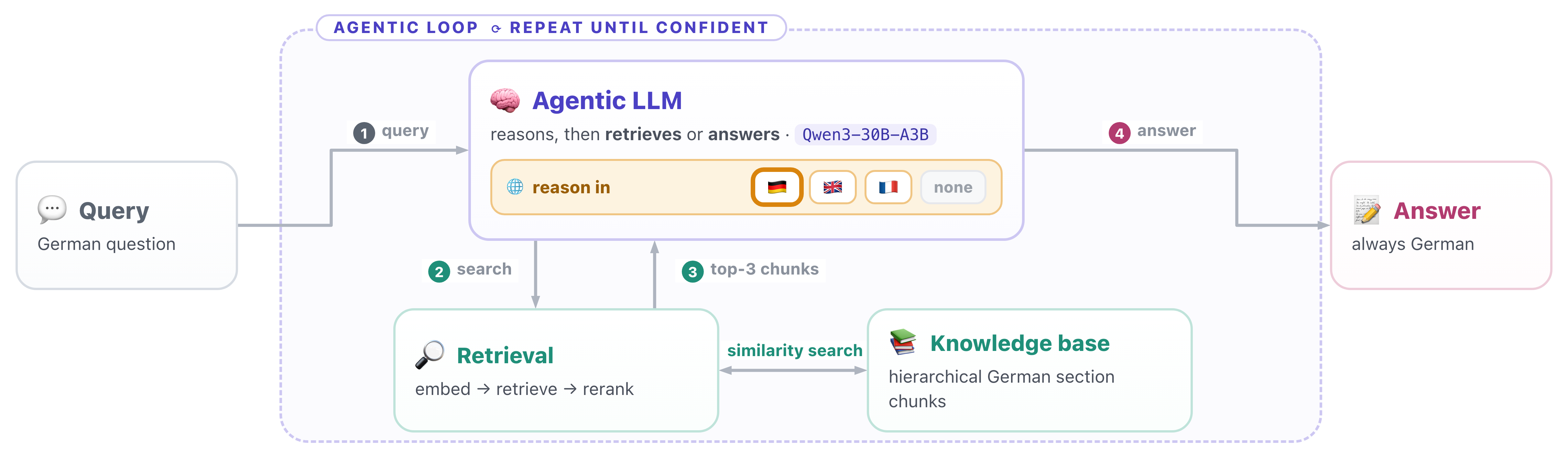}
\caption{Architecture of our agentic RAG system. Given a German query, the LLM reasons and then either issues a retrieval call or produces the final answer. Retrieval embeds the query, runs similarity search over the hierarchical German knowledge base, and reranks the hits to the top-3 chunks, which are appended to the context; the loop repeats until the model answers, always in German. The one control we vary is the language the model is forced to reason in (German, English, French, or none).}
\label{fig:arch}
\end{figure*}

The system is a standard agentic loop (Figure~\ref{fig:arch}). Given a query, the LLM is prompted with a system prompt, the query, and any evidence gathered so far, and it either (a) issues a retrieval tool call or (b) produces a final answer. The model has two research tools, both reading from the knowledge base: a \emph{RAG search} that returns the three most relevant chunks (top-30 retrieved, reranked to top-3), and a \emph{direct chunk retrieval} that fetches a chunk by its ID to walk to a parent or child section. A blacklist prevents already-seen chunks from being returned again for the same query. We implement inference, tool calling, and retrieval directly (without high-level agent frameworks) to keep full control of every step.

\paragraph{Controlling the reasoning language.}
By default the model reasons in English. To force another language, we follow the prompt-hacking method of \citet{qi2025reasoning}, where the system prompt instructs the model to reason in the target language and the reasoning trace is prefixed with a fixed short sentence in that language (e.g., German \emph{``Ich muss zunächst durchgehen, was ich über die Anfrage des Nutzers bereits weiß.''}\footnote{English translation: ``First, I need to go through what I already know about the user's request.''}). The final answer is always required to be in German. Prefixes for all languages are in Appendix~\ref{sec:appendix:prompts}.

\section{Experimental Setup}
\label{sec:setup}

\paragraph{Models and hyperparameters.}
All system runs use \texttt{Qwen3-30B-A3B}\footnote{Checkpoint: \href{https://huggingface.co/Qwen/Qwen3-30B-A3B}{Qwen/Qwen3-30B-A3B}} \citep{qwen3technicalreport} on a single 24\,GB GPU. As an additional no-retrieval baseline, we also run the larger \texttt{Qwen3-235B-A22B-Thinking}\footnote{Checkpoint: \href{https://huggingface.co/Qwen/Qwen3-235B-A22B-Thinking-2507}{Qwen/Qwen3-235B-A22B-Thinking}. We use this larger model only as a no-retrieval baseline and as the judge; running it inside the full RAG loop, with its repeated long-context tool calls, would be too expensive.} model from the same family through an API. We use the model's recommended decoding settings (\texttt{temperature}\,$=0.6$, \texttt{top\_p}\,$=0.95$, \texttt{top\_k}\,$=20$, and \texttt{min\_p}\,$=0$) and a fixed seed for reproducibility.

\paragraph{Retrieval.}
We embed chunks and queries with \texttt{Multilingual-E5-large}\footnote{Checkpoint: \href{https://huggingface.co/intfloat/multilingual-e5-large}{intfloat/multilingual-e5-large}} \citep{wang2024multilingual}. Retrieval is based on semantic similarity: we rank the chunks by the cosine similarity between the query embedding and each chunk embedding, and keep the 30 most similar chunks. We then rerank these 30 candidates with \texttt{Qwen3-Reranker-0.6B}\footnote{Checkpoint: \href{https://huggingface.co/Qwen/Qwen3-Reranker-0.6B}{Qwen/Qwen3-Reranker-0.6B}} \citep{qwen3embedding} and keep the top 3. We add this reranking step because the German and fictional data can be harder for the embedding model than typical retrieval data, so the first-stage similarity ranking is noisier. The reranker re-scores the top candidates and gives a more nuanced ordering.

\paragraph{Configurations.}
We compare: (i) \emph{NoRAG} baselines with the small and the large model and no retrieval; (ii) \emph{SimpleChunks}, a knowledge base of uniformly sized chunks; and (iii) \emph{SectionChunks}, our section-based knowledge base. For SimpleChunks we set the chunk size to the mean chunk size of SectionChunks so that both contain the same information at comparable granularity (Table~\ref{tab:chunkstats}). On both knowledge bases we vary the reasoning language: forced German, forced English, forced French, unconstrained (English, no prefix), and no reasoning (reasoning disabled). We use French as a control -- it is a high-resource language the model is not trained to reason in, and it differs from the German data, so it separates \emph{alignment with the data} from \emph{raw language ability}. SimpleChunks is run on the single-passage set with forced German, forced English, and unconstrained reasoning; SectionChunks is run with all five reasoning settings on both QA sets.

\paragraph{Metrics.}
Comparing free-form answers to a gold answer is hard, so we use two metrics. For the single-passage set we use LLM-as-a-Judge \citep{gu2025surveyllmjudge,li2025generationjudgment,ho2025llm}, where the larger \texttt{Qwen3-235B-A22B-Thinking} model, accessed through the same API, rates each answer from 1 to 5 given the question, the gold answer, and the source section (needed to tell extra correct detail from hallucination). For the multi-hop set, three human experts with deep knowledge of Aventuria rate answers from 1 to 5 under the same guidelines, blind to which configuration produced each answer. We validate the automatic metric against human ratings in Section~\ref{sec:results:judge}.

\begin{table}[t]
\centering
\small
\begin{tabular}{lrrrr}
\toprule
& \multicolumn{2}{c}{\textbf{SectionChunks}} & \multicolumn{2}{c}{\textbf{SimpleChunks}} \\
\cmidrule(lr){2-3}\cmidrule(lr){4-5}
\textbf{Book} & $C$ & $T_C$ & $\tilde{C}$ & $T_{\tilde{C}}$ \\
\midrule
G01 & 738  & 499 & 736  & 501 \\
G05 & 665  & 549 & 730  & 501 \\
G08 & 843  & 466 & 784  & 501 \\
\midrule
Total & 2246 & 501 & 2250 & 501 \\
\bottomrule
\end{tabular}
\caption{Knowledge-base statistics. $C$ is the number of chunks and $T_C$ the mean chunk size in tokens. SimpleChunks uses a fixed chunk size $T_{\tilde{C}}$ equal to the mean SectionChunks chunk size, so both hold the same 1.13M tokens in total at comparable granularity.}
\label{tab:chunkstats}
\end{table}

\section{Results and Discussion}
\label{sec:results}

\begin{figure}[t]
\centering
\includegraphics[width=\columnwidth]{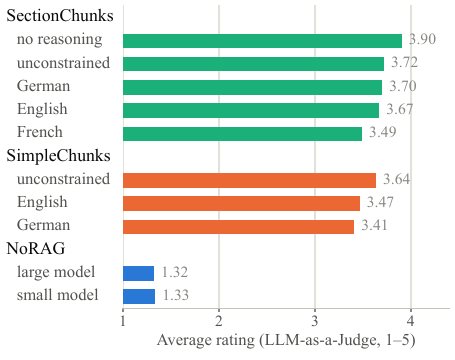}
\caption{Average ratings on the single-passage set. Retrieval (orange, green) is far above no-retrieval (blue). Among reasoning settings on the structure-aware SectionChunks base (green), forced German rises above forced French and reaches forced and unconstrained English, while disabling reasoning is highest.}
\label{fig:results}
\end{figure}

\begin{figure}[t]
\centering
\includegraphics[width=\columnwidth]{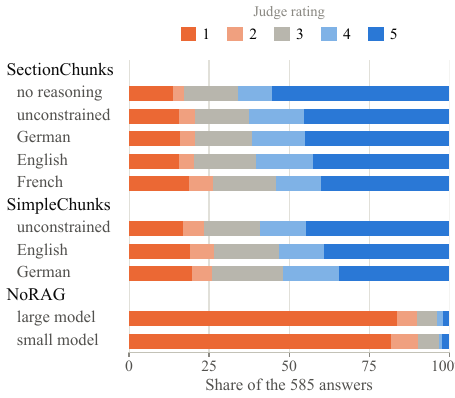}
\caption{Full rating distribution per configuration on the single-passage set (Table~\ref{tab:large} as shares). Colour encodes the judge rating from 1 (orange, hallucinated) to 5 (blue, correct). Retrieval moves the mass from rating~1 toward rating~5; for the retrieval settings rating~2 is the rarest.}
\label{fig:dist}
\end{figure}

Table~\ref{tab:large} reports the single-passage results (shown as averages in Figure~\ref{fig:results} and as a rating distribution in Figure~\ref{fig:dist}) and Table~\ref{tab:complex} reports the multi-hop results. Every configuration answers the same 585 questions, so we compare two configurations with the Wilcoxon signed-rank test \citep{wilcoxon1945} on their paired per-question ratings, and correct for multiple comparisons with the Holm method \citep{holm1979correction}. We report the resulting $p$-values below and give the full table in Appendix~\ref{sec:appendix:stats}.

\begin{table}[t]
\centering
\small
\setlength{\tabcolsep}{5pt}
\begin{tabular}{lrrrrrr}
\toprule
\textbf{Reasoning} & \textbf{1} & \textbf{2} & \textbf{3} & \textbf{4} & \textbf{5} & \textbf{Avg.} \\
\midrule
\multicolumn{7}{l}{\emph{NoRAG} (no retrieval)} \\
\quad small model    & 478 & 49 &  40 &   5 &  13 & 1.335 \\
\quad large model    & 490 & 36 &  36 &  11 &  12 & 1.323 \\
\midrule
\multicolumn{7}{l}{\emph{SimpleChunks} (fixed-size)} \\
\quad German         & 115 & 36 & 130 & 103 & 201 & 3.409 \\
\quad English        & 111 & 45 & 118 &  82 & 229 & 3.467 \\
\quad unconstrained  &  98 & 39 & 102 &  85 & 261 & 3.636 \\
\midrule
\multicolumn{7}{l}{\emph{SectionChunks} (structure-aware)} \\
\quad French         & 110 & 43 & 116 &  82 & 234 & 3.491 \\
\quad English        &  91 & 27 & 114 & 104 & 248 & 3.670 \\
\quad German         &  93 & 27 & 105 &  97 & 263 & 3.701 \\
\quad unconstrained  &  92 & 28 &  99 & 100 & 266 & 3.718 \\
\quad no reasoning   &  80 & 21 &  98 &  63 & 323 & \textbf{3.903} \\
\bottomrule
\end{tabular}
\caption{Single-passage set (585 questions), LLM-as-a-Judge ratings (1--5). Columns 1--5 give how many outputs received each rating; \emph{Avg.}\ is the mean. All questions and answers are in German; \emph{Reasoning} is the language the model is forced to reason in. Pairwise significance tests are in Appendix~\ref{sec:appendix:stats}.}
\label{tab:large}
\end{table}

\begin{table}[t]
\centering
\small
\begin{tabular}{lcccc}
\toprule
\textbf{Reasoning} & $R_A$ & $R_B$ & $R_C$ & \textbf{Avg.} \\
\midrule
NoRAG (large)   & 1.20 & 1.13 & 1.23 & 1.189 \\
\midrule
no reasoning    & 2.87 & 2.90 & 2.57 & 2.778 \\
French          & 2.87 & 2.70 & 2.77 & 2.778 \\
German          & 2.87 & 3.03 & 2.77 & 2.889 \\
English         & 3.30 & 2.83 & 2.87 & 3.000 \\
unconstrained   & 3.13 & 3.17 & 2.73 & \textbf{3.011} \\
\bottomrule
\end{tabular}
\caption{Multi-hop set (30 questions), human ratings (1--5) from three expert reviewers $R_A,R_B,R_C$. All use the SectionChunks knowledge base.}
\label{tab:complex}
\end{table}

\subsection{Retrieval Is Necessary}
Without retrieval both models score close to the minimum (1.335 and 1.323 on the single-passage set, 1.189 on the multi-hop set). Every retrieval configuration beats no retrieval by a large and significant margin ($p<10^{-70}$), and the two no-retrieval models are statistically tied ($p=.83$). The larger model does not help: it also hallucinates, only with entities that exist elsewhere in Aventuria rather than fully invented ones, and the judge penalizes both. The few correct no-retrieval answers concern facts that are not specific to Aventuria (for example, that mangroves take hold with stilt roots, which also holds in the real world). This confirms that the model does not know the domain and must use the retrieved German text -- exactly the property the testbed was designed for.

\subsection{Reasoning-language Alignment Helps, Independently of Proficiency}
\label{sec:results:align}
The clearest result is the comparison of forced German with forced French on SectionChunks. German scores 3.701 and French only 3.491, a gap of 0.21. This gap is statistically significant (Holm-corrected $p=.019$), and German wins on 182 questions against 138 for French. Both are forced, non-default reasoning languages, and both use the same data and system; the only difference is that German matches the language of the query and evidence while French does not. Crucially, according to the Qwen3 report, the model is \emph{more} proficient in French (average benchmark 71.8) than in German (68.2) \citep{qwen3technicalreport}. If the effect were about language ability, French should do at least as well as German. The gain therefore comes from \emph{alignment} with the language of the data, not from the model's language proficiency. The same ordering (German above French) also holds on the multi-hop set (2.889 vs.\ 2.778), so it is not an artifact of one evaluation.

\subsection{Richer Context Reduces the Cost of Forced Reasoning}
\label{sec:results:context}
Section-based chunks help every configuration over the fixed-size baseline, but not equally. Going from SimpleChunks to SectionChunks, unconstrained reasoning improves the least ($3.636\!\to\!3.718$, $\Delta=0.08$), and this gain is not significant ($p=.26$). Forced English improves more ($3.467\!\to\!3.670$, $\Delta=0.20$), and forced German improves the most ($3.409\!\to\!3.701$, $\Delta=0.29$), with a significant gain ($p<10^{-4}$). Two things follow. First, the general accuracy cost of forcing a reasoning language shrinks when the retrieved context is richer and carries German structure such as headings and metadata. With plain fixed-size chunks, forced German is the \emph{worst} of its group, but with section-based chunks it rises above forced English. Second, the configuration that benefits most from target-language structure is the one that also reasons in that language. This is consistent with our hypothesis that reasoning in the language of the evidence helps the model use that evidence and that the benefit is larger when there is more structured target-language context to reason over.
The two effects work together: forced German both reasons in German and has the most German structure to use, and it is the setting that gains most from richer context. 

\subsection{English Bias Remains}
While alignment narrows the gap, it does not fully close it. On the single-passage set, unconstrained English (3.718), forced German (3.701), and forced English (3.670) are statistically tied (all pairwise $p>.5$). So forced German reaches the level of the model's native, unconstrained English reasoning, but does not beat it. On the multi-hop set forced English (3.000) is slightly above forced German (2.889). In short, prompt-based language control lets German reasoning match unconstrained and forced English, but not surpass it. This matches the finding of \citet{qi2025reasoning} that prompt hacking gives only limited control. It suggests that closing the gap requires models with genuine multilingual reasoning (e.g., light fine-tuning to reason in the target language) rather than inference-time prompting alone. We see this as the main practical implication for RAG in lower-resource languages.

\subsection{When Does Reasoning Help?}
\label{sec:results:when}
On the single-passage set, \emph{disabling} reasoning is best of all (3.903), significantly above every reasoning setting on this set (Holm-corrected $p\le.04$). These questions are answerable from one retrieved passage, so a reasoning trace mostly adds noise and can pull in irrelevant content. On the multi-hop set the opposite happens, and no reasoning is now the \emph{worst} retrieval configuration (2.778), tied with French. This suggests reasoning helps only once a question needs synthesis across passages, and can hurt below that threshold. This adds a condition to prior results that reasoning improves multi-turn RAG \citep{yu2024autorag}, since the benefit depends on task complexity, which can plausibly be measured by the number of retrieval hops a question needs. In practice, this suggests that reasoning should be turned on based on the predicted difficulty of a question, rather than always on or always off. A RAG system could estimate how many hops a query needs and reason only above that threshold. Our two datasets show both cases, since reasoning hurts on one-hop questions and helps on multi-hop ones, but finding the exact threshold needs a benchmark graded by hop count, which we do not have.

\subsection{Judge Reliability and Language Artifacts}
\label{sec:results:judge}
To check the automatic metric, one author re-rated 50 answers (five per configuration) by hand and compared them to the judge; Table~\ref{tab:judge} shows the confusion matrix. Human and judge ratings agreed exactly in 66\% of cases (the diagonal) and within one point in 90\%; disagreements of two or more points occurred in 10\%. The metric is thus reliable enough to rank configurations but not to separate near-ties, which matters because our top configurations differ by little (e.g., unconstrained 3.718 vs.\ German 3.701). The larger disagreements are also \emph{asymmetric}. Four of the five gaps of two points or more are cases where the judge gave the minimum score of 1 to an answer a human rated 3 or 4. Since the judge is instructed to give a 1 whenever an answer appears to contradict the gold answer, it gives too low a rating to answers it considers hallucinated -- including, sometimes, correct ones (Appendix~\ref{sec:appendix:judgefail} shows a case where the system quotes the exact source sentence yet is scored 1).

\begin{table}[t]
\centering
\small
\setlength{\tabcolsep}{6pt}
\begin{tabular}{c ccccc}
\toprule
& \multicolumn{5}{c}{\textbf{Human rating}} \\
\cmidrule(lr){2-6}
\textbf{Judge} & 1 & 2 & 3 & 4 & 5 \\
\midrule
1 & \textbf{12} & 1 & 2 & 2 & 0 \\
2 & 0 & \textbf{1} & 0 & 0 & 0 \\
3 & 0 & 1 & \textbf{4} & 4 & 0 \\
4 & 0 & 1 & 0 & \textbf{2} & 2 \\
5 & 0 & 0 & 0 & 4 & \textbf{14} \\
\bottomrule
\end{tabular}
\caption{LLM-as-a-Judge vs.\ human ratings on 50 sampled answers (five per configuration). Exact agreement (\textbf{diagonal}) is 66\%; the larger disagreements sit in the top row, where the judge assigns 1 to an answer the human rates 3--4.}
\label{tab:judge}
\end{table}

\begin{figure}[t]
\centering
\includegraphics[width=\columnwidth]{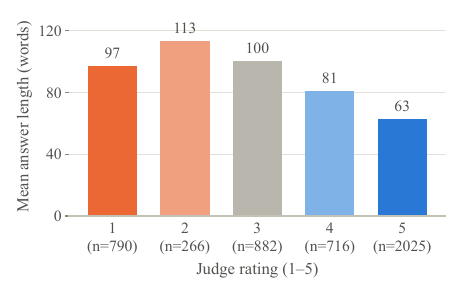}
\caption{Mean answer length by judge rating, pooled over the eight retrieval configurations on the single-passage set (4{,}680 graded answers). Higher-rated answers are shorter (Spearman $\rho=-0.39$, $p<0.001$); colour encodes the rating. This is direct evidence for the judge's length bias.}
\label{fig:answerlen}
\end{figure}

This asymmetry suggests another way to read one of our results. The judge punishes any apparent unsupported detail, so longer and more detailed answers give it more chances to flag something. Figure~\ref{fig:answerlen} confirms this. Pooled over all retrieval runs, answer length falls as the rating rises, from up to 113 words at the low ratings to 63 at rating~5 (Spearman $\rho=-0.39$, $p<0.001$). \emph{No reasoning} produces the shortest answers (64 words on average) and the highest ratings, so part of its advantage on the single-passage set (Section~\ref{sec:results:when}) is likely the judge rewarding short answers, rather than reasoning making the answers worse. However, this does \emph{not} affect our central alignment result. Forced German answers are slightly longer than forced French ones (85 vs 82 words), so the length bias works \emph{against} forced German, yet it still scores higher. The German-over-French ordering also holds on the multi-hop set, which is rated by humans with no judge involved. A second possible bias works in the opposite direction. The judge itself reasons in English and may overlook errors in non-English reasoning, which would favor the unconstrained and forced-English settings. A full human re-rating of the single-passage set is left to future work.

We also observe language artifacts. When not constrained to German, answers are more often grammatically flawed or contained non-German fragments (usually English, occasionally Chinese). More importantly for RAG, models forced to reason in French sometimes issue their \emph{retrieval queries} in French rather than German, querying a German knowledge base in the wrong language. This is a failure mode specific to \emph{agentic} RAG. Unlike the prompt-only setting of \citet{qi2025reasoning}, here the reasoning language can leak into the tool calls and hurt retrieval itself, not only the final answer. This makes the choice of reasoning language more important than in prompt-only reasoning. We report these as qualitative observations and leave quantifying them to future work.

\section{Conclusion}

We studied whether forcing an LLM to reason in the target language helps or hurts in a monolingual German RAG setting, where the query and the retrieved evidence are all in that language. Using a testbed over \dsa{} that the model cannot answer from memory, we found that aligning the reasoning language with the language of the data helps -- forced German beats forced French despite lower German proficiency -- and that this benefit grows when the retrieved context is richer and carries target-language structure, which also lowers the general cost of forcing a reasoning language. At the same time, forced German does not surpass the model's native, unconstrained English reasoning, so prompt-based control narrows but does not close the gap between English and other languages. Together, the results point to models with native multilingual reasoning when RAG is deployed over target-language corpora in lower-resource languages. We release the testbed and QA benchmark to support this line of work.

\section*{Limitations}

Our study is focused on purpose, and several limitations follow from that focus.

\paragraph{Single model and language pair.}
All system runs use one model family (Qwen3) and one language pair (German vs.\ English, with French as a control). Whether the effects transfer to other models and other non-English languages -- especially truly low-resource ones -- remains an open question.

\paragraph{Prompt-based control only.}
We control the reasoning language with prompt hacking, which is known to give only coarse control; we do not fine-tune the model to reason in the target language, which prior work suggests would help more \citep{qi2025reasoning}.

\paragraph{Small multi-hop set.}
The multi-hop set has 30 questions and shows non-trivial inter-reviewer disagreement, so its numbers are supplementary and should not be over-interpreted.

\paragraph{Imperfect judge.}
The LLM-as-a-Judge metric agrees with humans only within one point 90\% of the time and may be biased toward English reasoning; near-ties in Table~\ref{tab:large} are therefore uncertain.

\paragraph{Fixed retrieval.}
Retrieval hyperparameters were fixed (top-30 reranked to top-3), and the context length was limited by GPU memory. Trying a range of these retrieval settings would strengthen the conclusions.

\section*{Acknowledgments}
This work was supported by the Carl Zeiss Foundation through the MAINCE project (grant number P2022-08-009).

\bibliography{references}

\clearpage

\appendix

\section{PDF Processing Pipeline}
\label{sec:appendix:pipeline}

The knowledge base is built from region-description PDFs whose two- and three-column layout defeated the off-the-shelf converters we tried (Docling, Marker, pdfalto). Our semi-automatic pipeline converts each book in the following steps.

\begin{enumerate}
\item \textbf{PDF $\to$ ALTO XML.} We run pdfalto \citep{pdfalto}, which keeps per-token font and position information and the text-block structure. pdfalto also emits a stream of all parsed tokens, which we keep for later repair.
\item \textbf{ALTO $\to$ YAML.} We split the document by page and block, turn heading-font blocks into Markdown headings, and normalize bold/italic text. Words hyphenated across line breaks are rejoined using a corpus-wide word-count dictionary (and an LLM as a fallback) to decide whether the hyphen is real.
\item \textbf{Heading repair (semi-automatic).} Some words, including in headings, are dropped by pdfalto. We compare headings found via ALTO with those found by Marker on the same page; on a mismatch, the page is shown to a human, who assigns the correct heading level.
\item \textbf{Text repair (mostly automatic).} Missing words are recovered from the token stream by matching blocks with lowest Levenshtein distance. If more than 20\% of a block changes, the tool asks for confirmation.
\item \textbf{Reading-order repair (manual).} Because blocks in the PDF do not always follow the section order, a human reorders them (across page boundaries when needed) so that the section hierarchy is preserved.
\item \textbf{Load into chunks.} Each heading and its following text become a chunk; non-top-level chunks are linked as children of the nearest higher-level heading, giving the parent/child structure used at retrieval time.
\end{enumerate}

\section{Prompts and Reasoning Prefixes}
\label{sec:appendix:prompts}

To force the reasoning language, the system prompt is extended to require reasoning in the target language and the reasoning trace is prefixed with a fixed sentence:
\begin{itemize}
\item German: \emph{``Ich muss zunächst durchgehen, was ich über die Anfrage des Nutzers bereits weiß.''}
\item English: \emph{``First, I need to go through what I already know about the user's request.''}
\item French: \emph{``D'abord, je dois parcourir ce que je sais déjà sur la demande de l'utilisateur.''}
\end{itemize}
The final answer is always required to be in German. The LLM-as-a-Judge prompt provides the question, gold answer, system answer, and the source section, and asks for a single 1--5 rating; the human reviewers follow the same guidelines. When reading the rating, we kept the model's default tokenization, since how such short answers are tokenized can affect the result \citep{sanzguerrero2025mind}. For the local model, we use the default chat template and ensure that it contains no time-varying metadata (e.g., the current date) that could make our results vary \citep{sanzguerrero2026dating}. However, the judge and the large no-retrieval baseline run through a hosted API that injects the current date into the system prompt, so these results may not be fully reproducible. The multi-hop set is rated by humans and is not affected.

\section{A Judge Failure Case}
\label{sec:appendix:judgefail}

The judge assigns the minimum score whenever an answer looks unsupported, which occasionally misfires. In the case below (referenced in Section~\ref{sec:results:judge}) the question asks which kind of dragon nests in the coastal mountains by the Sea of the Seven Winds; the gold answer is \emph{Westwinddrachen}. The system answers exactly that and even quotes the source sentence, yet the judge scored it 1 while the human reviewer scored it 4.

\begin{center}
\small
\setlength{\fboxsep}{5pt}
\noindent\fbox{\begin{minipage}{0.93\columnwidth}
\raggedright
\textbf{Question.} \emph{Welche Drachenart nistet und jagt typischerweise in den Küstenbergen am Meer der Sieben Winde?}\\[3pt]
\textbf{Gold answer.} \emph{Westwinddrachen.}\\[3pt]
\textbf{System answer.} \emph{Die Drachenart, die \ldots{} nistet und jagt, ist die \textbf{Westwinddrache}. Dies ergibt sich aus dem Hinweis: ``In den Bergen und Klippen der Küsten am Meer der Sieben Winde [\ldots] horsten etliche Westwinddrachen.''}\\[3pt]
\textbf{Judge:} 1 \qquad \textbf{Human:} 4
\end{minipage}}
\end{center}

\begin{table*}[t]
\centering
\small
\begin{tabular}{lrcrr}
\toprule
\textbf{Comparison} & \textbf{$\Delta$} & \textbf{W/T/L} & \textbf{$p$} & \textbf{$p_{\mathrm{Holm}}$} \\
\midrule
no reason.\ $>$ French    & 0.41 & 214/263/108 & $8.2{\cdot}10^{-9}$ & $8.2{\cdot}10^{-8}$\,*** \\
no reason.\ $>$ English   & 0.23 & 195/265/125 & $.0008$ & $.007$\,** \\
unconstr.\ $>$ French     & 0.23 & 193/254/138 & $.0013$ & $.010$\,* \\
German $>$ French         & 0.21 & 182/265/138 & $.0027$ & $.019$\,* \\
no reason.\ $>$ German    & 0.20 & 190/270/125 & $.0033$ & $.020$\,* \\
no reason.\ $>$ unconstr. & 0.19 & 185/273/127 & $.0080$ & $.040$\,* \\
English $>$ French        & 0.18 & 186/248/151 & $.019$  & $.074$ \\
unconstr.\ $>$ English    & 0.05 & 185/245/155 & $.53$   & $1.00$ \\
German $>$ English        & 0.03 & 170/235/180 & $.56$   & $1.00$ \\
unconstr.\ $>$ German     & 0.02 & 173/250/162 & $.77$   & $1.00$ \\
\bottomrule
\end{tabular}
\caption{Pairwise Wilcoxon signed-rank tests on the single-passage set (585 paired ratings), for the five SectionChunks reasoning settings. $\Delta$ is the difference in mean rating; W/T/L counts questions where the first setting wins, ties, or loses; $p_{\text{Holm}}$ is corrected across the ten pairs. *** $p<.001$, ** $p<.01$, * $p<.05$.}
\label{tab:sig}
\end{table*}

The answer is correct and grounded in the retrieved text, but its added framing sentence appears to trigger the judge's contradiction check. Such cases are why the judge's low scores, in particular, should be read with care. The judge is itself an LLM, and LLMs are known to be poorly calibrated and often overconfident in their own outputs \citep{sanzguerrero2026overconfident}, so a confident minimum score is not by itself strong evidence that an answer is wrong. This points to a wider gap in how judges are validated: LLM-as-a-Judge assumes that the judge's confidence is meaningful, but validations -- including ours -- report agreement with humans and not calibration \citep{sanzguerrero2026calibration}. However, we find that the metric is still reliable enough to rank configurations (Section~\ref{sec:results:judge}).

\section{Statistical Significance}
\label{sec:appendix:stats}

All configurations answer the same 585 single-passage questions, so for any pair of configurations we have two ratings of the same question and can compare them question by question.

We use the Wilcoxon signed-rank test \citep{wilcoxon1945}. For each question it takes the difference between the two ratings, and it asks whether these differences lean toward one configuration or are just noise. It uses the direction and the relative size of the differences, not their exact values, so it fits ordinal 1--5 ratings and does not assume the ratings follow a normal distribution. It is the paired (same-question) counterpart of a $t$-test.

We run ten such tests, one for each pair of the five SectionChunks settings. Running many tests raises the chance that at least one looks significant by luck. The Holm correction \citep{holm1979correction} guards against this by making the threshold stricter the more tests we run, so that the chance of any false positive across all ten tests stays at 5\%. The corrected value $p_{\mathrm{Holm}}$ can be read directly against $0.05$. Table~\ref{tab:sig} reports, for each pair, the difference in mean rating ($\Delta$), how many questions the higher-mean configuration wins, ties, and loses (W/T/L), the raw $p$-value, and $p_{\mathrm{Holm}}$.

The three top reasoning settings -- unconstrained English, forced German, and forced English -- are not distinguishable from each other. Forced German is significantly above forced French, and disabling reasoning is significantly above all four reasoning settings.

The other tests reported in Section~\ref{sec:results} use the same Wilcoxon signed-rank test. Any retrieval configuration beats no retrieval with $p<10^{-70}$, and the small and large no-retrieval models are tied ($p=.83$). For the move from SimpleChunks to SectionChunks, we test each configuration's per-question gain (its SectionChunks rating minus its SimpleChunks rating). The gain for forced German is significant ($p<10^{-4}$), while the gain for unconstrained reasoning is not ($p=.26$).

\end{document}